\documentclass[preprint,3p,times,twocolumn,authoryear]{elsarticle}
\journal{Pattern Recognition Letters}

\usepackage{amsmath}
\usepackage{algorithmicx}
\usepackage{algpseudocode}
\usepackage{graphicx,xcolor}
\usepackage{placeins}

\usepackage{tikz}
\usepackage{mathdots}
\usepackage{color}
\usepackage{multirow}
\usepackage{tabularx}
\usepackage{extarrows}

\usepackage{booktabs}
\usepackage{tikz}
\usetikzlibrary{positioning, shapes, shadows, arrows}
\usetikzlibrary{decorations.pathreplacing}
\tikzstyle{abstract}=[circle, draw=black, fill=white]
\tikzstyle{labelnode}=[circle, draw=white,opacity=.2,text opacity=1]
\tikzstyle{invisiblenode}=[circle,dashed, inner sep=1pt,circle split,line width=1mm,minimum size=1.5cm]
\tikzstyle{line} = [draw, -latex']

\let\vec\mathbf

\renewcommand{\vec}[1]{\mathbf{#1}}

\begin{document}

\author[f2,ar]{Andrea Apicella} 

\affiliation[f2]{organization={Department of Electrical Engineering and Information Technology, University of Naples Federico II},
city={Naples},
country={Italy}}
\affiliation[ar]{organization={Laboratory of Augmented Reality for Health Monitoring (ARHeMLab)},
            city={Naples},
            country={Italy}}
\author[f2,ar]{Francesco Isgrò} 
\author[f2,ar]{Roberto Prevete}

\begin{frontmatter}

\title{Hidden Classification Layers: Enhancing linear separability between classes in neural networks layers}

\begin{abstract} 
In the context of classification problems, Deep Learning (DL) approaches represent state of art. Many DL approaches are based on variations of standard multi-layer feed-forward neural networks. These are also referred to as deep networks. The basic idea is that each hidden neural layer accomplishes a data transformation which is expected to make the data representation  ``somewhat more linearly separable"  than the previous one to obtain a final data representation which is as linearly separable as possible. However, determining the appropriate neural network parameters that can perform these transformations is a critical problem.  
In this paper, we investigate the impact on deep network classifier performances of a training approach favouring solutions where data representations at the hidden layers have a higher degree of linear separability between the classes with respect to standard methods. To this aim, we propose a neural network architecture which induces an error function involving the outputs of all the network layers. Although similar approaches have already been partially discussed in the past literature, here we propose a new architecture with a novel error function and an extensive experimental analysis. This experimental analysis was made in the context of image classification tasks considering four widely used datasets. The results show that our approach improves the accuracy on the test set in all the considered cases. 
\end{abstract}


\end{frontmatter}

\section{Introduction}
\let\thefootnote\relax\footnotetext{This paper has been published in its final version on \textit{Pattern Recognition Letters} journal in Open Access. The DOI is \url{https://doi.org/10.1016/j.patrec.2023.11.016}. Please refer to the peer-reviewed published version as main reference.}
\label{sec:introduction}
Nowadays, the success of Deep Learning (DL) approaches has led to an increase in interest in Multi-Layer Feed-Forward (MLFF) neural  networks \citep{lecun2015deep} insofar as a successful class of deep neural networks  consists  of MLFF networks with more than one hidden layer and possibly some specific architectural choices. In the rest of the paper we will refer to such Deep Neural Networks as DNNs.
In a nutshell, DNN networks are computational 
architectures organised as $L$ consecutive layers or levels of elementary computing units, called \textit{neurons}.
The last layer $L$ is the \textit{output} layer, and the remaining  layers are usually called \textit{hidden} or \textit{internal} layers.


In a DNN network each hidden layer $l$ performs a non-linear functional map $\Phi^l_{\mathbf{\theta}_l}$
from the output of the previous layer
$\mathbf{z}^{l-1}$ (and possibly other previous layers) to the output of the layer itself. 
Where $\mathbf{\theta}_l$ are the weights associated to the connections incoming into the layer $l$, plus the biases of the layer. By contrast, the output layer may also perform a linear transformation. 
In other words, the whole computation of a DNN can be viewed as a non-linear parametric functional mapping $\mathbf{y}= M(\mathbf{x};\mathbf{\theta})$ from a $d$-dimensional space to a $c$-dimensional space, where $d$ is the number of input variables and $c=m_L$ is the number of neurons in the output layer. 
The parameters $\mathbf{\theta}$ are the weights and biases of the network, and $\mathbf{y}$ are the output values of the output layer.

Although from a theoretical point of view the DNNs capability of being universal approximators has been extensively discussed \citep{cohen2016expressive,huang2000classification,longstaff1987pattern}, together with the inducted hidden feature representation spaces \citep{lerner1999comparative}, it is important to notice that the difficult to effectively find the most suitable $\mathbf{\theta}$ remains. In particular, when DNNs are applied in the context of classification problems, one has to find the parameters $\mathbf{\theta}_l$ such that the composition of $L-1$ non-linear transformations $\mathbf{z}^{L-1}=\Phi_{L-1}\Big(\Phi_{L-2}\big( \dots \Phi_1(\mathbf{x}) \big) \Big)$ maps each input $\mathbf{x}$ from a \textit{non-linearly} separable space into a \textit{linearly separable} one. In fact, in the context of classification problems, one of the main goals is to find a suitable data representation which allows to obtain a linearly separable classification problem. Plausibly, when a DNN is used, each internal representation $\mathbf{z}_{l}$ can be expected to make the representations of $\mathbf{x}$ ``somewhat more linearly separable''  than the previous one $\mathbf{z}_{l-1}$.  We underline that the complexity of a classification problem can be measured with respect different aspects, however class separability is a key aspect and different levels of class separability can be quantified \citep{lorena2019complex}.
In particular, in \citep{schilling2021quantifying} the Generalized Discrimination Value (GDV) to measure the separability between two dataset is introduced. The GDV is defined as the gap between the mean intra-cluster and the mean inter-cluster distances, computed on a set of labeled data represented in some space. More in detail, the GDV compares in a quantitative way the degree of class separability between two data representations. Since GDV can be computed on different types of representations, it can be also used to compare the separability of the same data represented in different spaces, such as the different representations returned by different neural networks' layers. 

However, we again emphasise that how to determine the appropriate parameters
$\mathbf{\theta}$ from a data set by a supervised learning process minimizing an error (or loss) function is still a critical problem. 
We notice that error functions usually depend on the final network output values only, without taking care about the results obtained in the hidden layers. Thus, starting from the previous considerations, in this paper we investigate the possibility to achieve a supervised learning approach which favours solutions where $\mathbf{x}$'s representations at the hidden levels have a higher degree of linear separability between the classes with respect to standard approaches. To this aim, we propose a DNN architecture which induces an error function involving the output values of all the network layers.
More specifically, as we will discuss in more detail in Section \ref{sec:method}, the output of each hidden layer $l$ is sent to an additional linear output layer which is trained to classify the input $x$ on the basis of the input representation encoding in the layer $l$ (see Figure \ref{fig:scheme}). From now on, we named this architecture Hidden Classification Layer network (HCL).
We investigated the impact of this type of solution in a series of experimental scenarios as we will discuss in more detail in Section \ref{sec:experim}.

Although similar approaches have already been partially discussed in the past literature (see, for example, \citep{lee2015deeply}), here we propose both a different version in terms of both neural architecture and error function, and a more extensive experimental analysis (see Sections \ref{sec:method} and \ref{sec:experim}). 
In particular, in \citep{lee2015deeply} the supervision of the hidden layer was made by SVMs instead of linear neural layers as in our case.  In \citep{wang2020cascade} a cascade of Convolutional Neural Networks (C-CNN) was proposed. C-CNN is composed of hidden layers combined together through dilated convolutions and trained using a proposed progress optimisation algorithm. Also in this case, our approach proposes a simpler architecture to favour hidden representations with a higher degree of class separability.
The rest of the paper is organised as follow: in Section  \ref{sec:method} the proposed method is described; Section \ref{sec:experim} describes the experimental setup and the evaluation methods; in Section \ref{sec:results} the results are reported and discussed; finally, Section \ref{sec:conclusion} contains final remarks.
\section{Model description}
\label{sec:method}

A neural network, as remined before, is structured in L layers of neurons. 
Each neuron $i$ belonging to the $l$-th layer, achieves a two-step computation (see \citep{bishop2006pattern}, chapter 4):  a linear combination $a^l_i$ of the neuron's inputs is computed first, and then the neuron output $z^l_i$ is computed by an activation function $f_l( \cdot )$, i.e., $z^l_i = f_l(a^l_i )$. Usually,  activation functions are non-linear function  (see \citep{apicella2021survey} for a review). 
The activation function input $a^l_i$ is usually computed on the basis of real values, said \textit{weights}, associated with the connections coming from the neurons belonging to the layer $l-1$ (and possibly from other previous layers) and a bias value associated to the neuron $i$. 
Each layer $l$ is composed of $m_l$ neurons, and the flow of computation proceeds from the the first hidden layer to the output layer in a forward-propagation fashion. 

In this research work, we focus on $C$-classes classification problems, with $C\geq 2$. In this context, Cross-Entropy (CE) loss \citep{wang2022comprehensive} is one of the most common loss function to be optimised. Given a dataset of N samples, DS=$\{ (\vec{x}^n,\vec{t}^n)\}_{n=1}^N$, CE for the n-th sample can be expressed as follows: 
$$CE^{(n)}(\vec{\theta}; \vec{y}^n,\vec{t}^n) = - \sum\limits_{c=1}^C t^n_{c}\log(y^n_c)$$

where $\vec{t}^n \in \{0,1\}^C$ is the one-hot encoding representation of the class label of the $n$-th sample of the dataset, and $\vec{y}^n=\vec{y}(\vec{x}^n; \vec{\theta})$ is the output of the neural network when it is fed with the input $\vec{x}^n$. Finally, $\vec{\theta}$ corresponds to all the network parameters. The total CE loss is equal to the sum of the single $CE^{(n)}$ over the dataset samples, i.e., $CE=\sum\limits_{n=1}^N CE^{(n)}$. As previously said, this loss formulation takes into account only the classification reported by the final layer of the network, without considering how the intermediate network levels affect the final classification scores. By contrast, in our model, HCL network, the data representation corresponding to the output of each hidden layer is used as input of a linear classifier so as to favour a data representation for each level as separable as possible. 

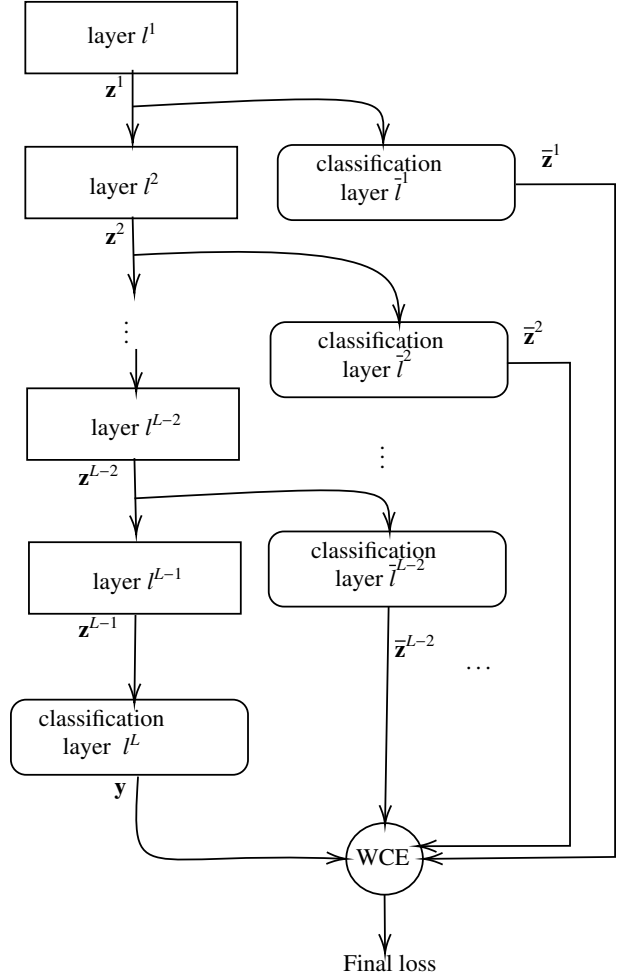
\begin{figure*}
    \centering
    \scalebox{0.9}{
    \tikzset{every picture/.style={line width=0.75pt}} 

\begin{tikzpicture}[x=0.75pt,y=0.75pt,yscale=-1,xscale=1]

\draw   (140,86) -- (257.5,86) -- (257.5,126) -- (140,126) -- cycle ;

\draw   (140,5) -- (257.5,5) -- (257.5,45) -- (140,45) -- cycle ;
\draw   (280,93.4) .. controls (280,88.76) and (283.76,85) .. (288.4,85) -- (403.1,85) .. controls (407.74,85) and (411.5,88.76) .. (411.5,93.4) -- (411.5,118.6) .. controls (411.5,123.24) and (407.74,127) .. (403.1,127) -- (288.4,127) .. controls (283.76,127) and (280,123.24) .. (280,118.6) -- cycle ;
\draw    (199.5,63.5) .. controls (344.78,55.7) and (338.9,57.88) .. (338.52,82.09) ;
\draw [shift={(338.5,84)}, rotate = 270] [color={rgb, 255:red, 0; green, 0; blue, 0 }  ][line width=0.75]    (10.93,-3.29) .. controls (6.95,-1.4) and (3.31,-0.3) .. (0,0) .. controls (3.31,0.3) and (6.95,1.4) .. (10.93,3.29)   ;
\draw    (199.5,43) -- (199.5,82) ;
\draw [shift={(199.5,84)}, rotate = 270] [color={rgb, 255:red, 0; green, 0; blue, 0 }  ][line width=0.75]    (10.93,-3.29) .. controls (6.95,-1.4) and (3.31,-0.3) .. (0,0) .. controls (3.31,0.3) and (6.95,1.4) .. (10.93,3.29)   ;
\draw    (200,146.5) .. controls (345.28,138.7) and (346.53,157.99) .. (347.43,183.06) ;
\draw [shift={(347.5,185)}, rotate = 267.8] [color={rgb, 255:red, 0; green, 0; blue, 0 }  ][line width=0.75]    (10.93,-3.29) .. controls (6.95,-1.4) and (3.31,-0.3) .. (0,0) .. controls (3.31,0.3) and (6.95,1.4) .. (10.93,3.29)   ;
\draw    (199.5,125) -- (200.45,166) ;
\draw [shift={(200.5,168)}, rotate = 268.67] [color={rgb, 255:red, 0; green, 0; blue, 0 }  ][line width=0.75]    (10.93,-3.29) .. controls (6.95,-1.4) and (3.31,-0.3) .. (0,0) .. controls (3.31,0.3) and (6.95,1.4) .. (10.93,3.29)   ;
\draw    (201.1,347.7) -- (200.52,395) ;
\draw [shift={(200.5,397)}, rotate = 270.7] [color={rgb, 255:red, 0; green, 0; blue, 0 }  ][line width=0.75]    (10.93,-3.29) .. controls (6.95,-1.4) and (3.31,-0.3) .. (0,0) .. controls (3.31,0.3) and (6.95,1.4) .. (10.93,3.29)   ;
\draw   (318,484) .. controls (318,472.95) and (327.51,464) .. (339.25,464) .. controls (350.99,464) and (360.5,472.95) .. (360.5,484) .. controls (360.5,495.05) and (350.99,504) .. (339.25,504) .. controls (327.51,504) and (318,495.05) .. (318,484) -- cycle ;
\draw    (202.5,438) .. controls (198.52,502.68) and (201.47,479.24) .. (316.26,483.93) ;
\draw [shift={(318,484)}, rotate = 182.46] [color={rgb, 255:red, 0; green, 0; blue, 0 }  ][line width=0.75]    (10.93,-3.29) .. controls (6.95,-1.4) and (3.31,-0.3) .. (0,0) .. controls (3.31,0.3) and (6.95,1.4) .. (10.93,3.29)   ;
\draw    (341.5,343) -- (339.53,463) ;
\draw [shift={(339.5,465)}, rotate = 270.94] [color={rgb, 255:red, 0; green, 0; blue, 0 }  ][line width=0.75]    (10.93,-3.29) .. controls (6.95,-1.4) and (3.31,-0.3) .. (0,0) .. controls (3.31,0.3) and (6.95,1.4) .. (10.93,3.29)   ;
\draw    (339.25,504) -- (339.48,535) ;
\draw [shift={(339.5,537)}, rotate = 269.57] [color={rgb, 255:red, 0; green, 0; blue, 0 }  ][line width=0.75]    (10.93,-3.29) .. controls (6.95,-1.4) and (3.31,-0.3) .. (0,0) .. controls (3.31,0.3) and (6.95,1.4) .. (10.93,3.29)   ;
\draw   (141,221) -- (258.5,221) -- (258.5,261) -- (141,261) -- cycle ;

\draw    (200.5,282.5) .. controls (345.78,274.7) and (341.42,275.17) .. (342.04,299.3) ;
\draw [shift={(342.1,301.2)}, rotate = 267.8] [color={rgb, 255:red, 0; green, 0; blue, 0 }  ][line width=0.75]    (10.93,-3.29) .. controls (6.95,-1.4) and (3.31,-0.3) .. (0,0) .. controls (3.31,0.3) and (6.95,1.4) .. (10.93,3.29)   ;
\draw    (200.5,260) -- (201.45,301) ;
\draw [shift={(201.5,303)}, rotate = 268.67] [color={rgb, 255:red, 0; green, 0; blue, 0 }  ][line width=0.75]    (10.93,-3.29) .. controls (6.95,-1.4) and (3.31,-0.3) .. (0,0) .. controls (3.31,0.3) and (6.95,1.4) .. (10.93,3.29)   ;
\draw   (142,307) -- (259.5,307) -- (259.5,347) -- (142,347) -- cycle ;

\draw    (201.5,199) -- (201.5,221) ;
\draw [shift={(201.5,223)}, rotate = 270] [color={rgb, 255:red, 0; green, 0; blue, 0 }  ][line width=0.75]    (10.93,-3.29) .. controls (6.95,-1.4) and (3.31,-0.3) .. (0,0) .. controls (3.31,0.3) and (6.95,1.4) .. (10.93,3.29)   ;
\draw   (276,192.4) .. controls (276,187.76) and (279.76,184) .. (284.4,184) -- (399.1,184) .. controls (403.74,184) and (407.5,187.76) .. (407.5,192.4) -- (407.5,217.6) .. controls (407.5,222.24) and (403.74,226) .. (399.1,226) -- (284.4,226) .. controls (279.76,226) and (276,222.24) .. (276,217.6) -- cycle ;
\draw   (275,309.4) .. controls (275,304.76) and (278.76,301) .. (283.4,301) -- (398.1,301) .. controls (402.74,301) and (406.5,304.76) .. (406.5,309.4) -- (406.5,334.6) .. controls (406.5,339.24) and (402.74,343) .. (398.1,343) -- (283.4,343) .. controls (278.76,343) and (275,339.24) .. (275,334.6) -- cycle ;
\draw   (132,403.4) .. controls (132,398.76) and (135.76,395) .. (140.4,395) -- (255.1,395) .. controls (259.74,395) and (263.5,398.76) .. (263.5,403.4) -- (263.5,428.6) .. controls (263.5,433.24) and (259.74,437) .. (255.1,437) -- (140.4,437) .. controls (135.76,437) and (132,433.24) .. (132,428.6) -- cycle ;

\draw    (412.1,107) -- (467.1,106.9) -- (467.1,482.9) -- (362.5,483.98) ;
\draw [shift={(360.5,484)}, rotate = 359.41] [color={rgb, 255:red, 0; green, 0; blue, 0 }  ][line width=0.75]    (10.93,-3.29) .. controls (6.95,-1.4) and (3.31,-0.3) .. (0,0) .. controls (3.31,0.3) and (6.95,1.4) .. (10.93,3.29)   ;
\draw    (407.1,206.9) -- (442.1,206.7) -- (442.1,476.7) -- (359.1,476.9) ;
\draw [shift={(357.1,476.9)}, rotate = 359.87] [color={rgb, 255:red, 0; green, 0; blue, 0 }  ][line width=0.75]    (10.93,-3.29) .. controls (6.95,-1.4) and (3.31,-0.3) .. (0,0) .. controls (3.31,0.3) and (6.95,1.4) .. (10.93,3.29)   ;

\draw (315,536) node [anchor=north west][inner sep=0.75pt]   [align=left] {Final loss};
\draw (173,15) node [anchor=north west][inner sep=0.75pt]   [align=left] {layer $\displaystyle l^{1}$ };
\draw (174,99) node [anchor=north west][inner sep=0.75pt]   [align=left] {layer $\displaystyle l^{2}$};
\draw (175,234) node [anchor=north west][inner sep=0.75pt]   [align=left] {layer $\displaystyle l^{L-2}$ };
\draw (193,174) node [anchor=north west][inner sep=0.75pt]   [align=left] {$\displaystyle \vdots $};
\draw (176,320) node [anchor=north west][inner sep=0.75pt]   [align=left] {layer $\displaystyle l^{L-1}$ };
\draw (136,399) node [anchor=north west][inner sep=0.75pt]   [align=left] { \ \ \ classification \\ \ \ \ \ \ \ \ layer \ $\displaystyle l^{L}$};
\draw (335,244) node [anchor=north west][inner sep=0.75pt]   [align=left] {$\displaystyle \vdots $};
\draw (406.94,373.94) node [anchor=north west][inner sep=0.75pt]  [rotate=-89.57] [align=left] {$\displaystyle \vdots $};
\draw (183,45) node [anchor=north west][inner sep=0.75pt]   [align=left] {$\displaystyle \mathbf{z}^{1}$};
\draw (183,127) node [anchor=north west][inner sep=0.75pt]   [align=left] {$\displaystyle \mathbf{z}^{2}$};
\draw (168,262) node [anchor=north west][inner sep=0.75pt]   [align=left] {$\displaystyle \mathbf{z}^{L-2}$};
\draw (168,347) node [anchor=north west][inner sep=0.75pt]   [align=left] {$\displaystyle \mathbf{z}^{L-1}$};
\draw (188,440) node [anchor=north west][inner sep=0.75pt]   [align=left] {$\displaystyle \mathbf{y}$};
\draw (425,83) node [anchor=north west][inner sep=0.75pt]   [align=left] {$\displaystyle \overline{\mathbf{z}}^{1}$};
\draw (415,182) node [anchor=north west][inner sep=0.75pt]   [align=left] {$\displaystyle \overline{\mathbf{z}}^{2}$};
\draw (343.5,357) node [anchor=north west][inner sep=0.75pt]   [align=left] {$\displaystyle \overline{\mathbf{z}}^{L-2}$};
\draw (300,89) node [anchor=north west][inner sep=0.75pt]   [align=left] {classification \\ \ \ \ \ layer $\displaystyle \overline{l}^{1}$};
\draw (301,188) node [anchor=north west][inner sep=0.75pt]   [align=left] {classification \\ \ \ \ \ layer $\displaystyle \overline{l}^{2}$};
\draw (297,305) node [anchor=north west][inner sep=0.75pt]   [align=left] {classification \\ \ \ \ \ layer $\displaystyle \overline{l}^{L-2}$};
\draw (322,477) node [anchor=north west][inner sep=0.75pt]   [align=left] {WCE};

\end{tikzpicture}
    }
    \caption{a scheme of the proposed approach. Each hidden layer $l^i$ of the main branch of the network produces an output $vec{z}^i$, and the final classification layer $l^L$ produces the output $vec{y}$. For each hidden layer $l^i,\ 1\leq i < L$, a further classification layer $\overline{l}^i$ is added. Each $\overline{l}^i$ is fed with the respective $\vec{z}^i$, producing an output $\overline{\vec{z}}^i$. Therefore, all outputs $\overline{\vec{z}}^i,$ $\forall 1\leq i < L$, are used together with the network classification output $\vec{y}$ to compute the final WCE loss.}
    \label{fig:scheme}
\end{figure*}

More formally, given a DNN composed of $l_1,l_2,\dots, l_{L-1}$ hidden layers and a final layer $l_{L}$ having $C$ neurons, 
we connect each hidden layer $l_j, \ 1 \leq j \leq L-2$ with a new layer $\overline{l}_j$ acting as an independent classifier. Adopting a proper loss function to train each classifier  $\overline{l}_j$, we expect that the features learned by the associate layer $l_j$ are the most discriminating as possible.
In other words, additional $L-2$ layers $\{\overline{l}_1, \overline{l}_2,\dots\overline{l}_{L-2}\}$ composed of $C$ neurons are added, and each $\overline{l}_j$ layer receives connections from the hidden layer $l_j$ only, 
making each $\overline{l}_j$ as an independent linear classifier. 
Therefore, given a DNN $M$, we obtain an HCL network $\overline{M}$ which will be composed of two distinct sets of layers: i) standard neural network layers $\{l_1,l_2, \dots, l_{L}\}$, composing M, and ii) \textit{hidden classification layers} $\{\overline{l}_1, \overline{l}_2, \dots, \overline{l}_{L-2}\}$, composing a set of layers  where each layer $\overline{l}_i$ favours more separable data representations in the respective hidden layer $l_i$, independently from the subsequent layers. Each hidden classification layer $\overline{l}_i$ has a set of parameters $\vec{\theta}^i$. Each $\vec{\theta}^i$ is composed of a set of distinct parameters, corresponding to the connections incoming in the layer $\overline{l}_i$, and a set of shared parameters with the other $\vec{\theta}^j$, with $j<i$, which correspond to the parameters of the DNN $M$ down to the layer $l_i$.   In figure \ref{fig:scheme} a general scheme of the proposed approach is reported.

Denoting with $\overline{\vec{z}}^{n,j}$ the scores returned by the hidden classification layer $\overline{l}_j$ on the $n$-th input sample, we propose the following Weighted Cross Entropy (WCE) loss formulation:
\begin{multline}
        WCE^{(n)}(\vec{\theta};\vec{y}^n,\vec{t}^n) = CE^{(n)}(\vec{\theta}_M;\vec{y}^{n},\vec{t}^{n}) + \\
+\sum\limits_{j=1}^{L-1} \lambda_j \cdot CE^{(n)}(\vec{\theta}^j; \overline{\vec{z}}^{n,j},\vec{t}^{n})
\end{multline}

where $\vec{y}^{n}$ is the score returned by the final layer of the classifier $M$, $\overline{\vec{z}^{n,j}}$ is the score returned by the hidden classification layer $\overline{l_{j}}$ tied to the $l_{j}$ layer, $\vec{\theta}^{M}$ are the parameters of the model $M$, and $\{\lambda_1, \lambda_2, \dots, \lambda_L\}$ is a set of regularisation coefficients greater than or equal to $0$.
Setting $\lambda_1=\lambda_2=\dots= \lambda_{L-1}=0$ results in standard CE loss applied to the final classification layer only, while different values give different weights to the hidden classification layers $\{\overline{l}_j\}_{j=1}^{L-1}$.

\section{Experimental assessment}
\label{sec:experim}
\subsection{Data and neural network models}
The performance of the HCM network architecture is assessed on image classification tasks considering four well-known datasets: \textit{MNIST}, \textit{Fashion MNIST}, \textit{CIFAR 10}, and \textit{CIFAR 100}. 
The MNIST\cite{lecun1998gradient} dataset consists of 70,000 grayscale images at a resolution of  $28 \times 28$ representing 10 different classes (the digits from $0$ to $9$). It is divided in two sets: the former composed of 60,000 images usually used as training samples and the latter of the remaining 10,000 images usually used as test samples. 
Fashion-MNIST is a dataset of images representing fashion articles \cite{xiao2017fashion}. Fashion-MNIST was proposed as a replacement for the original MNIST dataset for benchmarking machine learning algorithms, sharing the same image size and structure of training and testing splits. Indeed, it provides a training set of 60,000 examples and a test set of 10,000 examples. Each example is a 28x28 grayscale image, representing one of the following items: T-shirt/top, Trouser, Pullover, Dress, Coat, Sandal, Shirt, Sneaker, Bag, Ankle boot. 
CIFAR-10 dataset consists of 60,000 colour images of 10 different classes, that are \textit{airplane}, \textit{automobile}, \textit{bird}, \textit{cat}, \textit{deer}, \textit{dog}, \textit{frog}, \textit{horse}, \textit{ship}, and \textit{truck}. The dataset provides 50,000 training images and 10,000 test images at a resolution of $32 \times 32$.

Concerning the neural network models which are used as baseline to evaluate the classification enhancement of HCL network architecture, we considered 
LeNet-5 \cite{lecun1998gradient}, Hinton network \cite{hinton2012improving}, and ResNet18 \cite{he2016deep}. They are among the most famous networks exploiting convolutional layers for image classification. 

In its standard formulation, LeNet-5 is composed of a sequence of 3 convolutional layers interspersed  by 2 sub-sampling layers, followed by 2 final full-connected layers (the latest one for classification). 
Instead, Hinton network is composed of three convolutional hidden layers interleaved with three maxpooling layers. 
Finally, the main characteristic of a ResNet is the presence of shortcuts between non-consecutive layers that allows deep networks to be easily trained. In this work, we use the 18 layer residual network (ResNet18) described in \cite{he2016deep}.

\subsection{Evaluation}
Each model is evaluated on both the original version (\textit{vanilla}) proposed in their respective works and on its modified instance as described in Section \ref{sec:method}.  All the models were trained using the same experimental setup reported in their reference papers, except for Learning Rate $LR$, the number of Max Epochs $ME$, and the Patience Epochs $PE$, which can be strongly dependent by the network architecture. $ME$ and $PE$  are experimentally set to $1000$ and $200$ respectively, since we experimentally noticed that these values are enough to converge in all the analysed cases, while optimal $LR$  and $\lambda$ values are found through a grid-search approach. For the $LR$, the search space was $LR \in [10^{-5},10^{-1}]$, instead different combinations are considered for $\lambda$ parameters. Experiments on ResNet involving CIFAR 10 and CIFAR 100 dataset were made both considering only original data and augmented data, using $4$ pixels zero padding, corner cropping, and random flipping. 

Importantly, in order to experimentally show that the proposed method leads toward more easily separable data representations we computed  Generalized Discrimination Value (GDV) measure \cite{schilling2021quantifying} for each vanilla network's layer and its corresponding version equipped with hidden classification layers. We expect that, as the depth of the network increases, the data representations obtained with the proposed network's layout are more easily separable respect to the representations obtained by the respective models without additional layers.
Note that GDV is a measure of how well different data classes separate.
GDV values result $0.0$ for data points with randomly shuffled classes, and $-1.0$ in the case of perfectly separable classes. 
More in detail, GDV on a data representation $\mathbf{z}$ is defined as
\begin{multline*}
     GDV(\mathbf{z})=\frac{1}{\sqrt{D}}\Big(\frac{1}{L}\sum\limits_{c=1}^{C}d^{intra}(\mathbf{z}_c)+\\-\frac{2}{C(C-1)}\sum\limits_{c=1}^{C-1}\sum\limits_{m=c+1}^C d^{inter}(\mathbf{z}_c,\mathbf{z}_m)\Big)
\end{multline*}

where $d^{intra}(\mathbf{z}_c)$ is the mean intra-class distance on the data representations of the data $\mathbf{z}_c$ belonging to the class $c$ , and $d^{inter}(\mathbf{z}_c,\mathbf{z}_m)$ is the mean inter-class distance on the data representations  $\mathbf{z}_c,\mathbf{z}_m$ belonging to the $c$ and $m$ classes.


\section{Results}
\label{sec:results}
In Tab. \ref{tab:results} the test set accuracy, which was obtained by both the HCL network architecture and the vanilla networks, is reported. It is shown that the adoption of the HCL architecture improves the accuracy in all the cases, especially in the cases where a low accuracy for vanilla networks was obtained. In these cases, in fact, HCL network architecture appears to give a more significant improvement. In Fig. \ref{fig:gdv} the GDV values for each layer of each model and for CIFAR10 and CIFAR100 dataset are reported. 

\begin{table}[!htp]\centering

\scriptsize
\begin{tabular}{lrrrr}\toprule
&\textbf{Model} &\textbf{Baseline} &\textbf{Proposed} \\\midrule
\textbf{MNIST} &LeNet5 &99.0 &\textbf{99.2} \\
&Hinton &98.7 &\textbf{99.4} \\
\hline
\textbf{FMNIST} &LeNet5 &90.6 &\textbf{90.8} \\
&Hinton &92.1 &\textbf{92.6} \\
&ResNet &92.3 &\textbf{93.3} \\
\hline
\textbf{CIFAR10} &LeNet5 &66.2 &\textbf{71.5} \\
&Hinton &81.2 &\textbf{83.3} \\
&ResNet &86.3 &\textbf{89.0} \\
&ResNet (augmented) &94.4 &\textbf{94.6} \\
\hline
\textbf{CIFAR100} &LeNet5 &34.9 &\textbf{38.4} \\
&Hinton &52.5 &\textbf{53.3} \\
&ResNet &60.1 &\textbf{63.6} \\
&ResNet (augmented) &74.7 &\textbf{75.4} \\
\bottomrule
\end{tabular}
\caption{Results of the evaluation stage. For each dataset, accuracy on the test set obtained on both the vanilla models (\textit{baseline}) and the proposed ones are reported. ResNet has not been tested on MNIST dataset due to the already very high accuracies obtained with the other models. For CIFAR10 and CIFAR100, performance was evaluated also using augmented data (\textit{augmented}) for ResNet}
\label{tab:results}
\end{table}

\begin{figure*}
    \centering
    \includegraphics[width=1.1\textwidth]{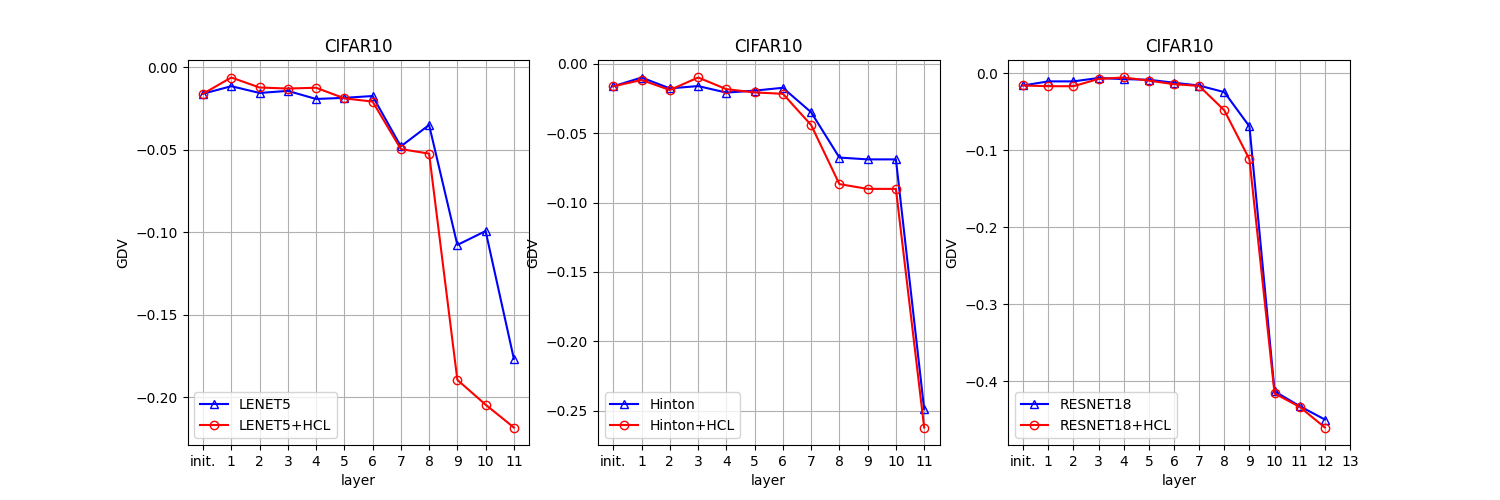}
    \caption{GDV values obtained with the proposed approach (LAT) compared with the vanilla networks on the CIFAR10 dataset. On the x axis, the layer of the model and on the y axis the respective GDV value.}
    \label{fig:gdv}
\end{figure*}

\begin{figure*}
    \centering
    \includegraphics[width=1.1\textwidth]{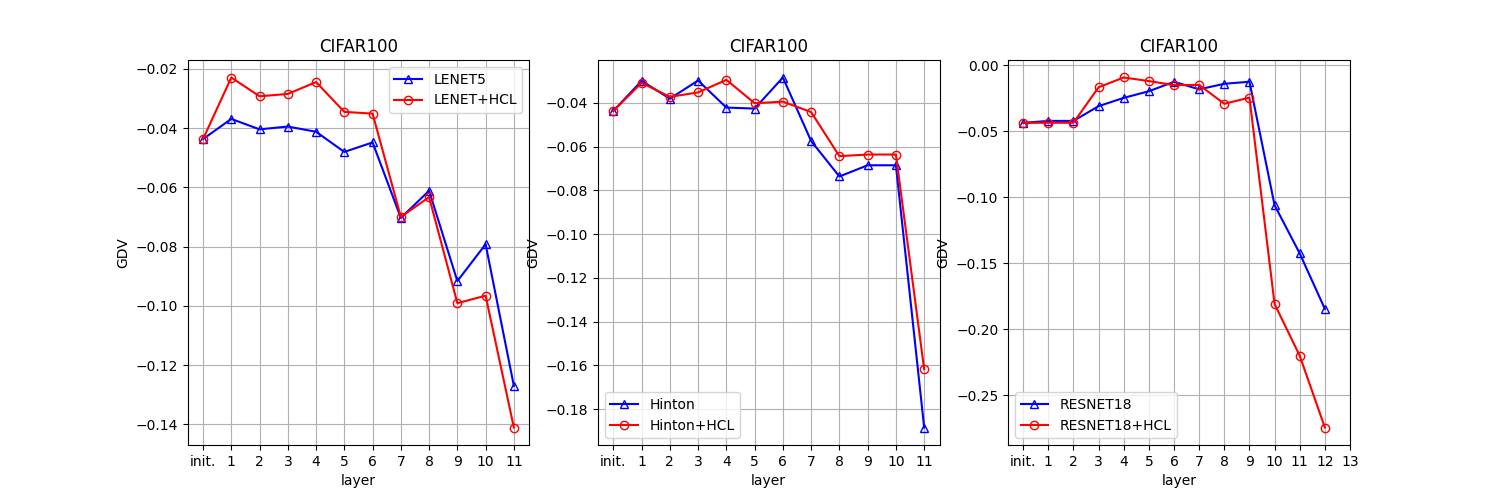}
    \caption{GDV values obtained with the proposed approach (LAT) compared with the vanilla networks on the CIFAR100 dataset. On the x axis, the layer of the model and on the y axis the respective GDV value.}
    \label{fig:gdv}
\end{figure*}
\section{Conclusion}
\label{sec:conclusion}
In this research work, we experimentally investigated the impact of constraining the classification complexity of the intermediate input representations with respect to their linear separability on the performances of DNNs in classification tasks. To this aim, we proposed a novel DNN architecture, which we named Hidden Classification Layer (HCL) network, where the output of each standard hidden layer is sent to a hidden classification layer trained to classify the input $x$ based on the $x$ representation given by the standard layer itself.
HCL network architecture allows obtaining solutions with input representations at the hidden levels having a lower classification complexity with respect to their linear separability. Note that our approach can be applied in slightly different ways: 1) given an already known neural network architecture, one can, first, augment it by hidden classification layers and, then, train the whole system from scratch; 2)  given an already trained neural network architecture, one can, first, augment it by hidden classification layers and, then, tune the whole system; 3) one can design and train a new neural architecture equipped with hidden classification layers.  
In this study, we used the first approach to test our proposal by considering three successful neural network models (LeNet-5, Hinton network, and ResNet18). These models were trained with and without hidden classification layers to evaluate the impact of HCL on the model performances experimentally. Each model was trained and tested on four datasets (MNIST, fashion-MNIST, CIFAR-10 and CIFAR-100). The results show that the HCL network has a positive impact uniformly (see Table \ref{tab:mnist_results,tab:fmnist_results,tab:cifar10_results,tab:cifar100_results}). 
It is interesting that the proposed approach leads to a GDV improvement in almost all cases, suggesting that the HCL network architecture can help the model build more separable inner representations. 
Moreover, it is worth noting that, in all the cases and with and without hidden classifier layers, the GDV values exhibit only a slight decrement for the initial network layers or they have even a wavering behaviour. Just only for the last layers, there is a sharp decrement (this decrease is particularly pronounced for HCL networks). 
Thus, these results are consistent with \citep{balduzzi2017shattered}, where the authors show experimentally that, during the learning phase, the loss derivatives with respect to the network parameters behave very similarly to a random walk on the first weight layers. In fact, in our case, data separability occurs mainly in the last layers of the network at the end of the learning process.

\section*{Funding}

This work  is supported by the European Union - FSE-REACT-EU, PON Research and Innovation 2014-2020 DM1062/2021 contract number 18-I-15350-2 and by the Ministry of University and Research, PRIN research project "BRIO – BIAS, RISK, OPACITY in AI: design, verification and development of Trustworthy AI.", Project no. 2020SSKZ7R .

\bibliographystyle{elsarticle-num-names} 
\bibliography{references}
\end{document}